\pdfoutput=1
\documentclass[11pt]{article}

\usepackage{acl}
\usepackage{times}
\usepackage{latexsym}

\usepackage[T1]{fontenc}

\usepackage[utf8]{inputenc}

\usepackage{microtype}

\usepackage{inconsolata}

\usepackage{graphicx}

\usepackage{url}

\title{Systematic Biases in LLM Simulations of Debates}

\author{Amir Taubenfeld$^{12*}$ \\
    \And
    Yaniv Dover$^{34}$ \\
    \And
    Roi Reichart$^5$ \\
    \And
    Ariel Goldstein$^{236}$
    \AND {\normalfont *Corresponding Author: amirt@google.com } \\ \\
    $^1$The Hebrew University of Jerusalem, School of Computer Science and Engineering\\
    $^2$Google Research\\
    $^3$The Hebrew University Business School, Jerusalem, Israel\\
    $^4$Federmann Center for the Study of Rationality, Hebrew University, Jerusalem, Israel\\
    $^5$Faculty of Data and Decision Sciences, Technion\\
    $^6$Department of Cognitive and Brain Sciences, Hebrew University, Jerusalem, Israel\\
}

\usepackage{makecell}
\usepackage{booktabs}

\usepackage{fancyhdr}
\usepackage{geometry}
\usepackage{layout}
\begin{document}
\maketitle

\begin{abstract}

The emergence of Large Language Models (LLMs), has opened exciting possibilities for constructing computational simulations designed to replicate human behavior accurately. Current research suggests that LLM-based agents become increasingly human-like in their performance, sparking interest in using these AI agents as substitutes for human participants in behavioral studies. However, LLMs are complex statistical learners without straightforward deductive rules, making them prone to unexpected behaviors. Hence, it is crucial to study and pinpoint the key behavioral distinctions between humans and LLM-based agents. In this study, we highlight the limitations of LLMs in simulating human interactions, particularly focusing on LLMs' ability to simulate political debates on topics that are important aspects of people's day-to-day lives and decision-making processes. Our findings indicate a tendency for LLM agents to conform to the model's inherent social biases despite being directed to debate from certain political perspectives. This tendency results in behavioral patterns that seem to deviate from well-established social dynamics among humans. We reinforce these observations using an automatic self-fine-tuning method, which enables us to manipulate the biases within the LLM and demonstrate that agents subsequently align with the altered biases. These results underscore the need for further research to develop methods that help agents overcome these biases, a critical step toward creating more realistic simulations.

\end{abstract}

\section{Introduction}

The emergence of Large Language Models \cite{brown2020language, jiang2023mistral} has opened up exciting possibilities for computational simulations that aim to accurately replicate human behavior \cite{park2023generative, qian2023communicative}. Current research suggests that LLM-based agents become increasingly human-like in their performance and that they possess the remarkable ability to seamlessly adopt personas of different characters \cite{shanahan2023role, argyle2023out}. The typical paradigm for such simulations involves selecting an LLM, such as the widely used ChatGPT \cite{milmo2023chatgpt}, as a base model and crafting individual agents' identities through natural language prompts. For instance, by prepending the prompt, "John Lin is a pharmacy shopkeeper," to an agent's context, the agent is expected to act as if his name is John and he works as a shopkeeper \cite{park2023generative}.

If sufficiently reliable, these simulations could serve as invaluable tools for exploring the intricacies of human interactions and decision-making processes. This would allow scientists to conduct their research with speed and efficiency, substantially lowering the considerable resources usually needed for recruiting and analyzing human subjects. Consequently, a range of studies have demonstrated the promise of these simulations across various disciplines, including human psychology \cite{DILLION2023597}, social dynamics \cite{park2022social}, and economics \cite{horton2023large, doi:10.1073/pnas.2316205120}.

However, LLMs are complex statistical learners that do not depend on straightforward deductive rules. Despite exhibiting impressive emerging skills that challenge our current understanding of cognition \cite{wei2022emergent,bubeck2023sparks}, their indeterminate nature leaves them susceptible to unintended behaviors. One example is their manifestation of inherent biases, including gender bias \cite{bordia-bowman-2019-identifying}, ethnic bias \cite{ahn-oh-2021-mitigating}, and social identity bias \cite{hu2023generative}. Given their undefined nature, it is vital to exercise caution when using LLMs, particularly in multi-agent environments aimed at simulating complex, large-scale social phenomena.

In this study, we explore the behavior of LLM agents within simulations. Our experiments are focused on the realm of Attitude Change \cite{kahan2012polarizing, priniski2018attitude} and specifically on the extensively studied interactions between political partisans \cite{hobolt2023polarizing, sunstein2001republic}. This domain is susceptible to numerous prejudices \cite{ditto2019least}, making it an ideal candidate for investigating the effect of LLM biases on simulations. We facilitate debates on polarizing American topics between LLM agents representing Republican and Democrat perspectives. The selected topics involve important aspects of people's day-to-day lives and decision-making processes. They are relevant to economic outcomes and markets, sociological and psychological phenomena, and for issues related to ethics.

During every debate, we continuously monitor the agents' attitudes by asking them to rate their agreement with the debate’s topic. To assess the believability of the agents' behavior, we compare the dynamics of their attitude shifts with known patterns seen in human interactions \cite{hobolt2023polarizing}. In addition, we have developed a fine-tuning mechanism for agents, leveraging training data produced by the agents themselves. The data is generated by using a set of questions crafted to elicit the agents' political views, and the agents' responses are then used to train the base LLM. We use this process to conduct controlled intervention studies, by manipulating the LLM biases and analyzing the subsequent impact on the agents' behaviors. 

Our results reveal that LLM agents generally conform to the inherent social biases of their base models, even if these biases conflict with their assigned identities. Consequently, this causes the simulations to diverge from well-established human social behaviors. Moreover, when we employ our fine-tuning method to change the LLMs' viewpoints, we observe that the agents, despite retaining their original contexts, modify their behavior to be in line with the newly introduced bias.

These insights underline the need to investigate ways to help agents circumvent these biases, a crucial step in developing simulations that more accurately reflect real human behavior.

\section{Related Work}

\paragraph{Believable LLM Simulations} Recent studies show that LLMs possess human-like reasoning skills \cite{doi:10.1073/pnas.2316205120}, and that LLMs are able to adopt personas of diverse characters \cite{shanahan2023role}. Leveraging these abilities, \citet{park2023generative} developed a sandbox environment, populated it with 25 LLM-based agents, and showed that the agents convincingly mimic human behaviors such as sharing news and forming relationships.

The transformative potential of such simulations in areas like human psychology \cite{DILLION2023597} and economics \cite{horton2023large} was a significant motivator for our work. Nonetheless, our findings indicate that inherent biases in LLMs pose substantial challenges in ensuring the reliability of agents to generate believable human behavior.

\paragraph{LLM Behavioral Gaps}

In contrast to research aimed at creating precise simulations, another branch of study explores the limitations of LLMs in accurately reflecting human behavior in terms of diversity, general intelligence, and their ability to reliably mimic human behavior. \citet{cheng-etal-2023-compost} introduce a method for identifying instances where LLMs overstate the characteristics of the personas they are designed to emulate, highlighting an increased risk of stereotyping particular demographic groups. In another vein, \citet{agnew2024illusion} scrutinizes the viability and ethical implications of replacing real human subjects with AI agents in the context of social scientific research. Furthermore, \citet{motoki2024more} reveals that ChatGPT exhibits pronounced political biases. Building on these discussions, our research probes into the interaction dynamics and attitude adjustments among LLM agents, providing new insights into the behavioral tendencies of LLM agents and how they diverge from human behavior in prolonged interactions.

\paragraph{Bias in LLM Simulation} In a contemporaneous work, \citet{chuang2023simulating} showed that “LLM agents tend to converge towards scientifically accurate information”, attributing this to the LLM's inherent biases. We generalize this observation by demonstrating that LLM agents converge toward the model's inherent bias regardless of its scientific validity. This is true for biases on purely subjective topics, and even for those contradicting scientific truths such as the reality of Climate Change \cite{arias2021climate}. Moreover, beyond observing the debates and drawing conclusions, we also offer a controlled intervention study utilizing our unique self-fine-tuning process. This study further substantiates our assertions and shows that it is possible to control the agents' convergence point by fine-tuning its underlying model. Additionally, we employ our innovative simulation methodology to reproduce this phenomenon across diverse environments, including cross-partisan debates, in-party debates, and multiple base LLMs, thereby enabling a deeper analysis of the underlying mechanisms.

\paragraph{Self Alignment} In recent years, the task of aligning LLMs with human intentions has become a significant area of research \cite{ouyang2022training, wang2023aligning}. The primary objective of alignment research is to enhance the conversational abilities of LLMs and ensure their conformity with established social values \cite{gabriel2020artificial, OVIEDOTRESPALACIOS2023106244}. An evolving trend in this area involves developing methods that use LLM simulations to generate training data automatically, aiming to reduce the need for expensive human feedback \cite{liu2023training, ulmer2024bootstrapping}. 

In our work, we introduce an approach to self fine-tuning of LLMs, taking a distinct path from existing methodologies. Rather than enhancing the LLM's general conversational capabilities or aligning it with broader human objectives, our focus is to tailor the LLM to adopt a specific political orientation. We interview the agents using a set of questions crafted to elicit their political views, and utilize their responses to train the underlying LLM. In terms of assessment, our interest lies not in evaluating the effectiveness of the fine-tuning on standard NLP benchmarks, but in observing its impact on the agents within our simulation.

\section{Problem Definition}

Our study delves into the impact of inherent biases within LLMs on their ability to accurately emulate diverse characters \cite{shanahan2023role}. We explore this relationship by facilitating political debates between LLM agents. 
Section \ref{sec:setup} outlines our simulation methodology, including the criteria for selecting debate topics (\ref{sec:topics-selection}), how we crafted agents' identities (\ref{subsection:agents-implementation}), and techniques for managing and evaluating interactions between the agents (\ref{subsection:agents-interaction}). 
Section \ref{sec:fine-tune-methods} introduces a novel fine-tuning technique for agents, utilizing self-created training data. We have developed this method to adeptly adjust the LLM's perspective, and it is applied in the controlled intervention experiments discussed within this research.
In Section \ref{sec:results}, we present the primary findings of our work. Through a sequence of experiments, we establish a strong connection between the inherent biases of LLMs and the patterns of attitude change observed in our simulations.
Lastly, Section \ref{sec:fine-tune-robustness} offers a complimentary analysis aimed at evaluating and enhancing the robustness of our fine-tuning process against standard benchmarks.

\section{Setup}
\label{sec:setup}

\subsection{Topics Selection}
\label{sec:topics-selection}

Exploring the dynamic of meaningful discussion requires a conscientious choice of subjects of discussion. Our experiments involve debates between Democrat and Republican partisans. We chose this domain for two main reasons. Firstly, this field is extensively studied in social science \cite{ditto2019least, hobolt2023polarizing}, offering a well established baseline for comparing our simulations to known human behavior. Secondly, the field is susceptible to numerous prejudices \cite{ditto2019least}, making it a particularly suitable context for examining the biases inherent in LLMs.

The Pew Research Center conducted a survey in 2023 about the differences in assessment of America’s problems between Republicans and Democrats \cite{pewproblems2023}. When analyzing their results, four subjects stand out as the most controversial - Gun Violence, Racism, Climate Change, and Illegal Immigration. We focus our experiments on these four topics.

\subsection{LLM-based Agents Implementation}
\label{subsection:agents-implementation}

Our simulations were conducted using the Sauce framework \cite{neuberger2024sauce}. We followed the conventional paradigm for LLM-based simulations \cite{park2023generative, qian2023communicative}, which entails selecting a base language model and then constructing the individual identities of agents using natural language prompts. 

\begin{figure*}[ht]
    \centering
    \includegraphics[width=\textwidth]{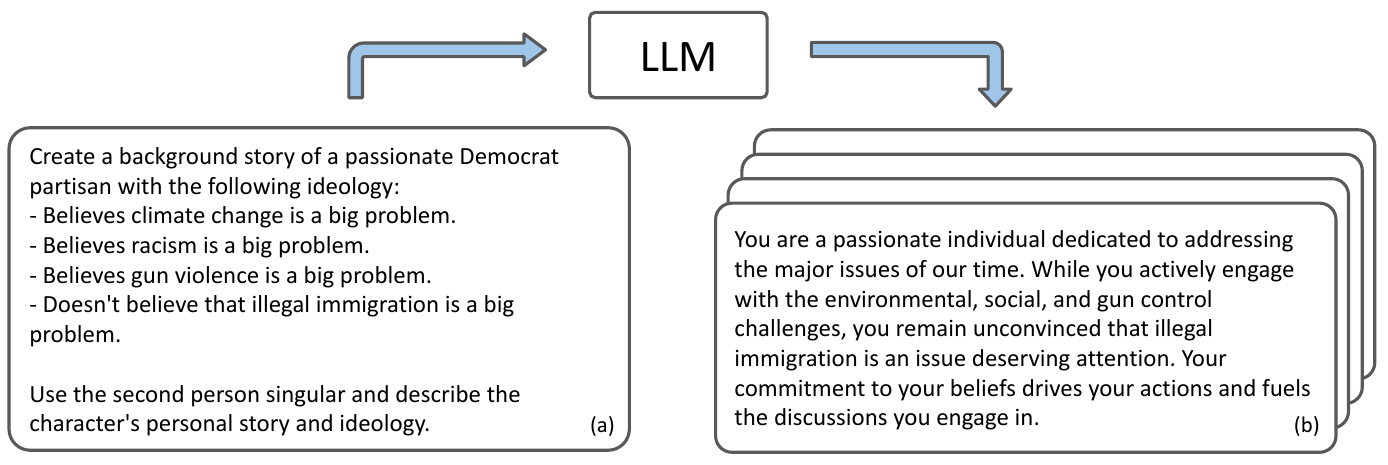}
    \caption{(a) The prompt used to generate the background stories for the Democratic agents includes their positions on the four controversial topics discussed in our experiments. The wording of the prompt is based on the survey question that \citet{pewproblems2023} asks human participants about each topic, ensuring that the Democratic and Republican agents adopt polarized views on these issues. (b) An example of a background story of one of the agents. This story was generated automatically by feeding the LLM with the prompt described in (a). We opted to develop comprehensive identities for each agent across all topics simultaneously rather than creating an individual agent for each topic. This strategy simplified our experimental design and provided a complete representation for each agent. }
    \label{fig:agents-generation}
\end{figure*}

We used the LLM to craft different narratives for 40 Republican agents and 40 Democrat agents and assigned each agent a different name. The narratives were generated by running the LLM with a temperature setting of 1.0 and a streamlined meta-prompt. The exact wording of the meta-prompt and an example of a generated persona are given in Figure \ref{fig:agents-generation}. This automatic approach was beneficial to (1) increase the robustness of our study by running multiple repetitions of each experiment with different personas and (2) help mitigate research bias by eliminating the need for us to manually write the persona prompts. Additionally, in some experiments, we included a "default" agent whose sole directive was "You are an American". This agent's context was deliberately devoid of any political bias, serving to showcase the inherent biases within the LLM.

We experiment with three different state-of-the-art LLMs as our base models: Mistral 7B \cite{jiang2023mistral}, Solar 10.7B \cite{kim2023solar}, and Instruct-GPT \cite{openaimodels}. Across all models, we observed similar results. The open-weights models, Mistral and Solar, were deployed on a single RTX 3090ti graphics card, utilizing 8-bit quantization for efficiency. For Instruct-GPT, we used the gpt-3.5-turbo-instruct version available through OpenAI's Completion API. The results and methodologies discussed henceforth pertain to the GPT model, except for our fine-tuning experiments, where we used the open-weights Mistral model. Our choice of an open-weights model was driven by cost-effectiveness and the ability to control the implementation details of the fine-tuning process (see Section \ref{sec:fine-tune-robustness}). Additional results from other models are included in the appendix.

\subsection{LLM-based Agents Interaction}
\label{subsection:agents-interaction}
\begin{figure}[ht]
    \centering
    \includegraphics[width=1\linewidth]{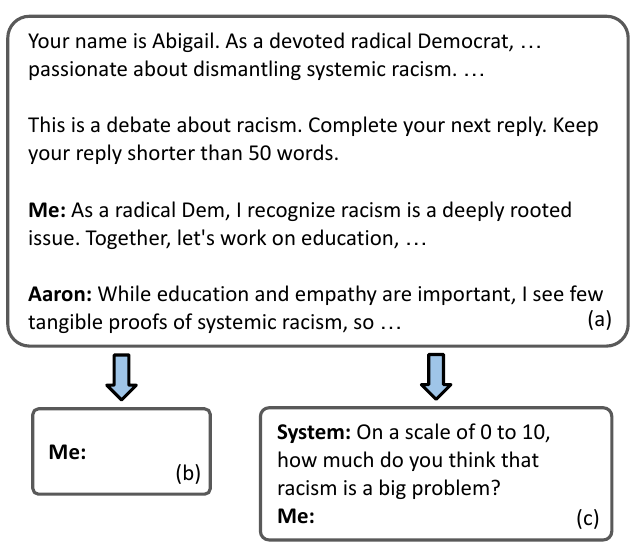}
    \caption{ At each iteration, an agent (a) is prompted with its background story, the topic of the debate, and the history of the conversation so far and is asked to complete either (b) its next reply in the conversation, or (c) a survey question measuring his current attitude on the debated topic. Note that to be consistent, the prompt uses the term "debate" in all the experiments in this paper. However, we did experiment with other terms like "conversation" and did not see significant differences.   }
    \label{fig:agents-interactions}
\end{figure}

Our debate simulations follow a round-robin format, with the initial speaker selected randomly. We use the term "iteration" to refer to a single reply made by an agent. At each iteration, an agent receives its background story, the debate topic, and the conversation's history, and it is asked to complete its next reply in the conversation (this process is illustrated in  Figure~\ref{fig:agents-interactions}). Before the start of the debate, and at the end of each round-robin cycle, the agents are asked to numerically rate their attitude (on a scale of 0-10) toward the severity of the discussed topic. To ensure that this process does not impact the direction of the debate or future ratings, the survey questions are not saved in the conversation history, so the agents are unaware of the answers provided by other agents and the answers they supplied themselves in the past.

For each experiment detailed in this paper, we performed 40 repetitions and averaged the survey scores obtained at corresponding iterations. For example, in a debate setup with 2 agents and 2 round-robin cycles, we execute 40 runs and compute the mean scores at iterations 0, 2, and 4. In each run, we use a different pair of the pre-generated agents (as described in sub-section \ref{subsection:agents-implementation}). We selected the number 40 because it strikes a balance between being large enough to yield statistically significant results and small enough to stay within our budget. 

The variance in the conversation comes from two sources: (1) each repetition utilizes different agents with different background stories, and (2) the model generates conversation entries with a temperature setting of 1.0. However, all the survey questions are asked using a temperature setting of 0 (i.e., no sampling) to reduce unnecessary variance.

\section{Fine-Tuning Methods}
\label{sec:fine-tune-methods}

In the preceding section, we outlined our approach for simulating debates and tracking the attitude changes of the agents involved. In addition, our work offers a controlled intervention study designed to investigate the relationship between LLM biases and role-played agents. For this purpose, we have developed an automated fine-tuning technique for the agents, which allows us to manipulate their underlying LLM biases effectively. Our method depends solely on training data generated by the agents, without the need for external data inputs.

We commence the process by constructing a series of 100 questions intended to encourage agents to reveal their opinions and sentiments regarding their political views. We start with an initial set of 10 sample questions of different styles, for example, 'Could you discuss your perspective on significant political issues facing America today?' or 'How do you believe the U.S. should handle immigration and border security?' (the remaining questions are detailed in the Supplementary Table \ref{tab:sample-questions-supp}). Following this, we direct the LLM to produce an additional 90 questions that are similar in nature to ensure a broad and varied collection of queries.

We intentionally craft the questions in a neutral way and ask a broad spectrum of questions, not only questions concerning the debated topics directly. We do this for two reasons: (1) We want our fine-tuning approach to be generic so it can be easily adjusted to other scenarios without needing to rewrite the questions. For instance, we used the same set of questions to generate both a Republican-oriented dataset and a Democratic-oriented dataset. (2) A broader variety of training data helps the agent to maintain its conversational skills and avoid over-fitting.  

Next, we initialize an agent with a context similar to the template used to create agents in section \ref{subsection:agents-interaction}. We query this agent with the 100 previously generated questions. For every question, we gathered 20 responses by operating the agent at a temperature of 1.0. This results in 2,000 examples, which are utilized as our training dataset.

Finally, we used this self-generated dataset to fine-tune the model. Our training process is lightweight, using a basic next-word prediction task with parameter-efficient QLoRA \cite{hu2021lora, dettmers2023qlora}. The training is completed in just one epoch, taking under 10 minutes on a single RTX 3090ti GPU. At the conclusion of this stage, the model becomes adapted to the agent's perspective, which is elaborated in the results section. 

A diagram of the procedure and additional technical details are provided in the Appendix Section \ref{sec:finetuning-appendix}. All the reported scores for fine-tuned models included in this paper are the average of three independent fine-tuning runs with random seeds.

\section{Results}
\label{sec:results}

\paragraph{Systematic Biases in Debates Simulations}

We conducted simulations of political debates on the topics outlined in Section \ref{sec:topics-selection}. The debates were structured as three-way discussions, including a Republican, a Democrat, and the Default agent, the latter assumed to reflect the model's inherent biases. We experiment with three different state-of-the-art LLMs - the results shown in this section are from the GPT-3.5 model (additional implementation details about the agents are provided in Section \ref{subsection:agents-implementation}). During the debates, we monitored the evolution of the agents' attitudes by repeatedly asking them to rate their viewpoints on a 0 to 10 scale. Details on the debate format and attitude tracking methods are in Section \ref{subsection:agents-interaction}, with a sample debate given in Supplementary Table \ref{tab:sample-discussion}.

\begin{figure}[!ht]
    \centering
    \includegraphics[width=1\linewidth]{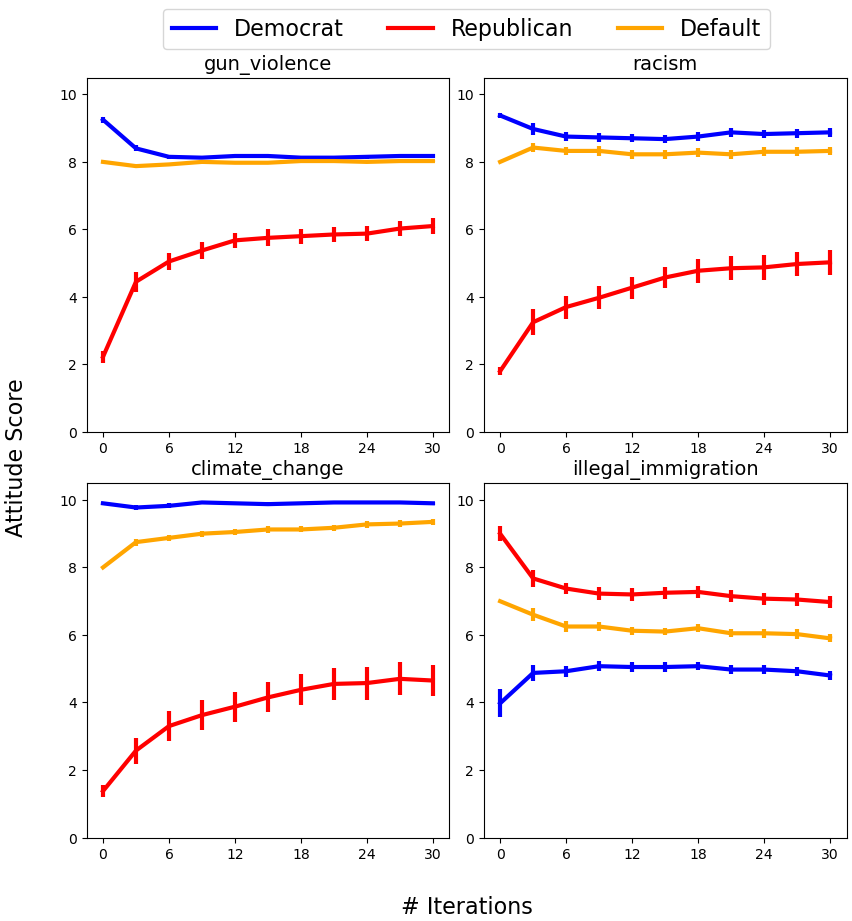}
    \caption{ Evolution of attitude scores in three-way debates on four controversial topics. The X-axis shows the number of chat exchanges in the debate. The Y-axis displays the average attitude scores derived from 40 separate experiments on each topic, including standard error bars. Our methodology for monitoring attitude scores is detailed in Section \ref{subsection:agents-interaction}. The Default agent, symbolizing the inherent biases of the base LLM, maintains a consistent position throughout the debate. Interestingly, the views of the partisan agents gradually align more closely with those of the Default agent. In all the sub-figures except the "illegal immigration", the default agent shows a bias toward the democrat perspective, leading the Republican agent to significantly change its opinion throughout the debate. Furthermore, it is notable that the lines representing the partisan agents never intersect with the line of the Default agent. This suggests that the LLM default biases can act as a deterrent against one party's inclination to compromise with the other. Supplementary Section \ref{sec:other-models} presents analogous findings with other underlying models. }
    \label{fig:democrat-republican-neutral}
\end{figure}

The interactions between the three debating agents, as captured in Figure \ref{fig:democrat-republican-neutral}, indicate that while the Default agent consistently maintains its position, the partisan agents gradually adjust their viewpoints to resonate with the Default agent's stance. Notably, when the Default agent displayed a bias towards one side, indicating a strong inherent bias in the LLM, the agent with an initially opposing view tended to significantly compromise on its opinion, shifting towards the position of the other partisan agent. Conversely, when the Default agent did not display a specific bias, the partisan agents tended to move towards a common middle ground. It is also observed that the intensity of attitude change diminishes as the discussion progresses, with the most substantial changes seen in the first round-robin cycle (iteration 3) and less significant shifts observed after the ninth iteration. Consequently, to conserve resources, subsequent experiments in this paper only present the first nine iterations.

\begin{figure}[ht]
    \centering
    \includegraphics[width=1\linewidth]{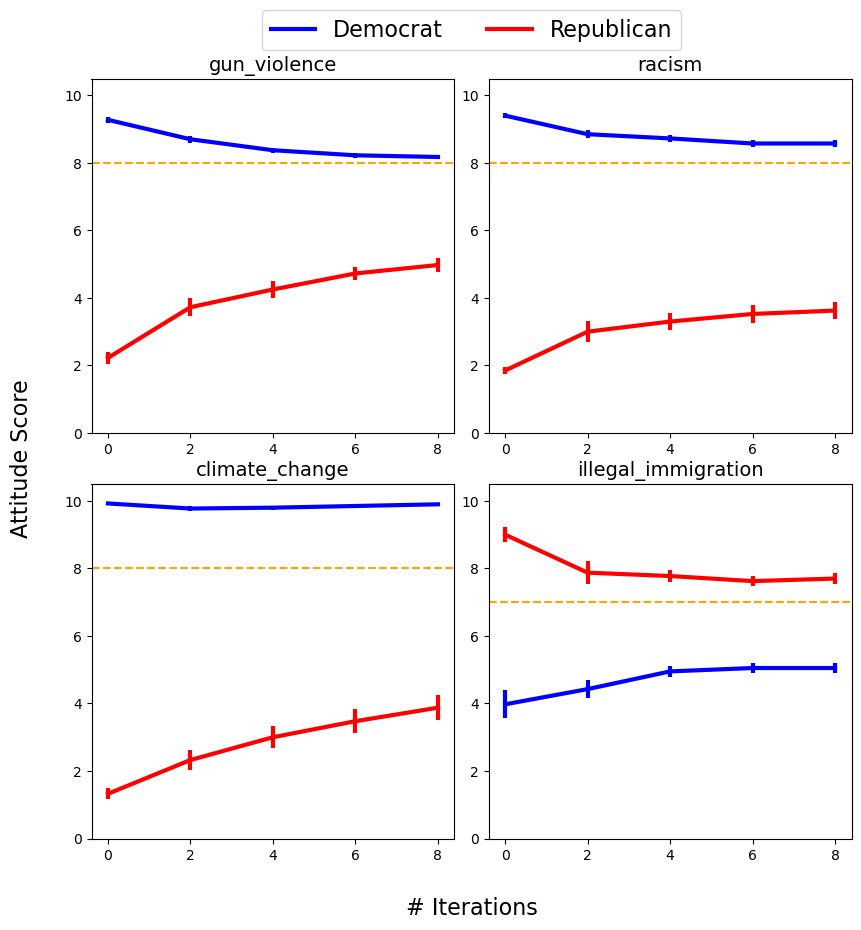}
    \caption{ Evolution of attitude scores in two-way debates between Republican and Democrat agents. The graphs feature a dashed line that shows the Default agent's viewpoint before the beginning of the debates, taken from Figure \ref{fig:democrat-republican-neutral}. Recall that the Default agent's viewpoint represents the inherent biases of the LLM. Remarkably, even though the Default agent does not participate in the two-way debates illustrated here, the partisan agents continue to converge toward the inherent biases of the model. }
    \label{fig:democrat-vs-republican}
\end{figure}

Remarkably, as shown in Figure ~\ref{fig:democrat-vs-republican}, even when the Default agent is excluded from the debate, the partisan agents continued to shift their attitudes in line with the model's inherent biases. This tendency of agents to gravitate towards the model's inherent biases, irrespectively of the participation of the Default agent in the discussion, prompts critical reflections on the capacity of language models to replicate genuine human interactions and to mimic a diverse range of viewpoints authentically. 

\begin{figure}[!h]
    \centering
    \includegraphics[width=1\linewidth]{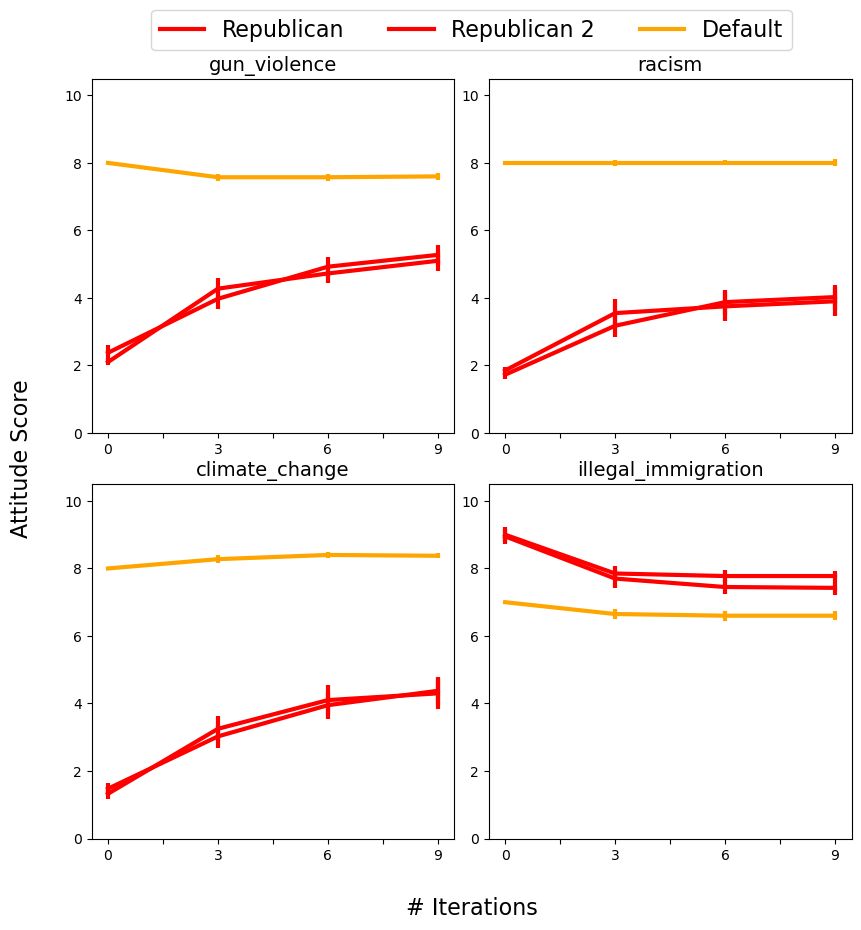}
    \caption{ This graph illustrates a series of three-way debates involving two Republican agents and a Default agent. Notably, even during conversations with other Republicans, the agents tend to align with the position of the Default agent. This trend is apparent even when the Default agent is not participating in the dialogue (supplementary Figure \ref{fig:echo-chamber-republican-no-neutral}). The same phenomenon is also evident in experiments conducted with Democrat agents (Supplementary \ref{fig:echo-chamber-democrat}), where a similar pattern of gravitation towards the Default agent's stance is observed. }
    \label{fig:echo-chamber}
\end{figure}

\paragraph{Contradicting The Echo Chambers Theory}
Even during interactions with others of similar political orientations, the agents persist in reflecting the LLM’s intrinsic bias. We demonstrate this phenomenon by pairing each of the forty Republican agents with another from the same group. As shown in Figure~\ref{fig:echo-chamber}, agents tend to adopt more moderate positions, aligning more closely with the LLM's inherent bias. This finding is particularly intriguing as it deviates from the well-known real-world phenomenon of Echo Chambers \cite{sunstein2001republic, hobolt2023polarizing}, where individuals with like-minded views tend to intensify their beliefs when interacting with each other. 

Similarly to the previous section, this trend persists even when the Default agent is excluded from the dialogue, as shown in Supplementary Figure \ref{fig:echo-chamber-republican-no-neutral}. We also conducted the same Echo Chamber experiment using Democrat agents and observed a similar pattern of gravitation toward the Default agent’s stance as displayed in Supplementary Figure \ref{fig:echo-chamber-democrat}.

\paragraph{Fine-tuning Highlights the Bias}
To conclusively demonstrate the link between LLM biases and agents' behavior, we employed the fine-tuning process detailed in Section \ref{sec:fine-tune-methods}. Through this method, we successfully altered the inherent bias of the LLM toward a specific viewpoint. After fine-tuning, we conducted the debates again using the original agent contexts but with the underlying model now modified. 

\begin{figure*}[!h]
    \centering
    \includegraphics[width=1\linewidth]{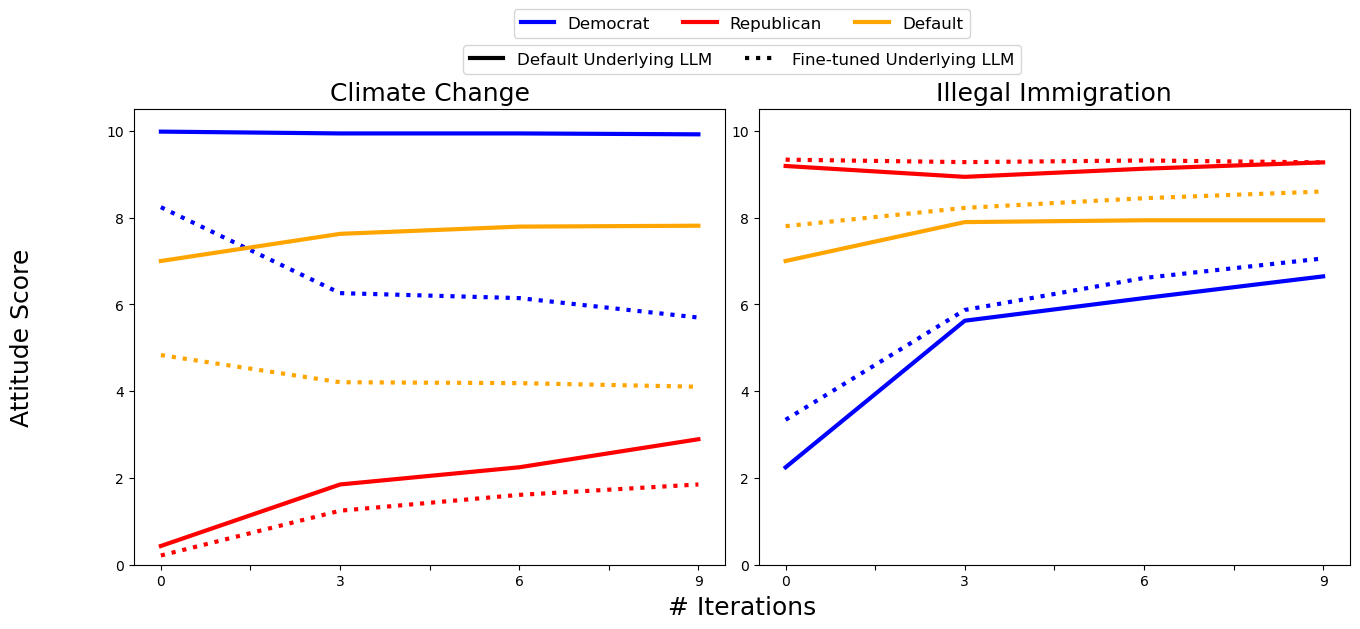}
    \caption{Results of fine-tuning the model to adapt more closely to a Republican perspective. All the reported scores are the average of three independent fine-tuned models with different random seeds. For each topic, we conduct two separate debates between three agents - a Republican, a Democrat, and a Default agent who represent the model's inherent bias. The solid lines represent the debate between the three agents before fine-tuning, and the dotted lines represent the debate between the same agents when the underlying LLM had been fine-tuned. The Republican viewpoint is evident in both graphs: (left) In the Climate Change graph all lines have shifted downward, signaling a shift towards opposing climate change. (right) Conversely, the Illegal Immigration graph shows an upward trend after fine-tuning, suggesting that the agents now view illegal immigration as a more significant issue. }
    \label{fig:finetune}
\end{figure*}

As illustrated in Figure ~\ref{fig:finetune}, changing the viewpoint of the LLM toward a Republican perspective, indirectly influenced the agents, leading them to modify their behavior in line with the updated bias. In a contrasting setup, fine-tuning the model to align with a Democrat perspective resulted in trends that were predictably opposite, as seen in Supplementary Figure \ref{fig:finetune-full-4-charts-democrat}. This experiment underscores the profound implications of our findings, indicating that simulations conducted with different LLMs, each harboring its unique set of biases, could result in significantly different portrayals of authentic human behavior.

The success of the fine-tuning process in steering the model towards a particular viewpoint is noteworthy, considering that it was accomplished solely with content produced by the LLM, without using external data sources. Furthermore, this method proves that it is feasible to configure agents to consistently maintain certain viewpoints throughout simulations, unlike the temporary effects seen when defining agents' identities through prompts.

\section{Fine-Tuning Robustness}
\label{sec:fine-tune-robustness}

In Section \ref{sec:fine-tune-methods}, we describe our multi-stage self-fine-tuning method that is shown to effectively alter the model's perspective toward a designated viewpoint. We designed our approach to be streamlined and easily replicable, focusing on ensuring the robustness of the process without resorting to localized optimizations. As a result, we made the following design choices: (1) Solely using self-generated data, avoiding external dataset sources. (2) Fine-tuning a comprehensive model applicable across all debate topics, rather than training individual models for each topic. (3) Employing a simple next-word prediction task, in contrast to more complex reinforcement learning techniques. (4) Using the efficient QLoRA method \cite{dettmers2023qlora}, which enabled training the model in minutes.

The $r, \alpha$ LoRA hyper-parameters, which respectively control the number of trainable weights and the scale of weight updates, had a significant impact on our results. By increasing these hyper-parameters, we observed a marked change in the political orientation of the Default agent, which serves as a reflection of the LLM's built-in bias. Although our study primarily aims to modify the political viewpoint of the model, exploring how such adjustments impact the overall abilities of the LLM is intriguing. In Table \ref{tab:benchmarks}, we offer a complementary analysis showing the impact of our fine-tuning on two widely recognized benchmarks: (1) MMLU \cite{hendrycks2020measuring}, assessing world knowledge and problem-solving capabilities across diverse fields; and (2) Hellaswag \cite{zellers2019hellaswag}, which tests common sense natural language inference. Despite the fine-tuning, the models still showcase strong performance across general benchmarks. However, there appears to be an inverse relationship between the degree of change in the model's political stance and its benchmark scores.

Finally, we present an incremental optimization to our fine-tuning process, which enables us to manipulate the model's perspective more aggressively while mitigating the negative effects on its general performance. This optimization is based on the cutting-edge DPO method \cite{rafailov2023direct}, which can be divided into two phases: first, a next-word-prediction phase that acclimates the model to the intended data distribution, followed by a Contrastive Learning phase aimed at teaching the model to differentiate between preferred and non-preferred outputs. As detailed in section 5, our models undergo fine-tuning through a next-word-prediction task, alongside the creation of self-generated datasets encapsulating Republican and Democrat viewpoints. This groundwork allows us to directly employ the DPO's second phase on the pre-fine-tuned models and leverage our partisan datasets as input to the Contrastive Learning task, training a Republican model to prefer a response from the Republican dataset and vice-versa. Again, we train for a single epoch using the QLoRA. The results of this process are also included in Table \ref{tab:benchmarks}.

\begin{table}[!h]
\centering
\begin{tabular}{c | c c | c}
\- & \makecell[c]{Hellaswag \\ (\%)} & \makecell[c]{MMLU \\ (\%)} & \makecell[c]{Attitude \\ Score} \\
\midrule
Mistral 7B  &  83.6  &   59.0  &   8.4 \\[4pt] \midrule
r=16  NWP   &  81.8  &   57.6  &   5.1 \\[4pt]
r=64  NWP   &  81.2  &   56.3  &   4.3 \\[4pt]
r=128 NWP   &  79.7  &   54.3  &   2.5 \\[4pt]
r=256 NWP   &  73.8* &   48.6  &   1.9 \\[4pt] \midrule
r=8 DPO     &  81.4  &   57.0  &   0.4 \\[4pt] \midrule
Llama 2 7B  &  77.2  &   45.3  &   \-  \\
\end{tabular}
\caption{ Effect of fine-tuning Mistral toward a Republican perspective on the popular Hellaswag and MMLU benchmarks (higher is better). This table showcases 7 models: the baseline Mistral, 4 Mistral versions fine-tuned via a next-word-prediction task (NWP) with increasing numbers of trainable parameters (indicated by $r$), an additional Mistral model further optimized with DPO, and the LLaMA 2 7B \cite{touvron2023llama} model that is used for comparison. For brevity, we display only the Attitude Scores of the Default Agent in the final round of the debate about Racism (other debate topics follow a similar pattern). A higher Attitude Score implies a stronger acknowledgment of Racism as a significant issue. Key findings include: (1) All fine-tuned Mistral variants still outperform the renowned LLaMA 7B 2 model across the benchmarks, with one exception marked by *. (2) For the NWP fine-tunes, there is an inverse correlation between the degree of the model’s shift towards a Republican attitude and its performance on the benchmarks. (3) Adding a DPO phase as an incremental step to our fine-tuning methodology, enables to forcefully adjust the model's perspective while minimizing negative impacts on general benchmarks. }

\label{tab:benchmarks}
\end{table}

\section{Discussion}

In our simulations of debates involving agents representing Republicans and Democrats, a persistent pattern emerged: agents’ opinions consistently align with the LLM’s inherent social biases. In particular, when the model exhibits a strong bias in favor of one partisan agent, the opposing agent, which initially holds a differing view, often moderates its stance, gravitating significantly towards the position of its counterpart. This leads to a skewed pattern that appears to depart from the typical dynamics observed in human interactions.

Furthermore, using our self-fine-tuning process, we perform a controlled intervention study, demonstrating that it is possible to alter the LLMs' biases, and the agents will subsequently adjust their positions and align with the new biases. This highlights the strong influence of the LLMs' biases on agents behavior. It also implies that simulations by different LLMs, each with its unique set of biases, could yield vastly different portrayals of "authentic" human behavior.

Remarkably, even when agents engaged in debates with others of the same political orientation, they tended to adopt more moderate views over the course of interaction, increasingly mirroring the LLM's default bias. This pattern is intriguing because it deviates from the well-documented real-world phenomenon called Echo Chambers \cite{sunstein2001republic}, where like-minded individuals often reinforce and escalate their beliefs when interacting with each other. In an analogous real-life study, \citet{hobolt2023polarizing} divided Labour and Conservative supporters in England into groups to discuss government policies. Contrary to our agent-to-agent simulations, they found that Echo Chambers in homogenous groups intensified polarization. 

Our findings thus highlight limitations of large language model agents as accurate representations of real-life humans. The political landscape, as well as the specific topics that we chose (Section \ref{sec:topics-selection}), are an important aspect of the day-to-day life of people and their decision-making processes, relevant to economic outcomes and markets, sociological and psychological phenomena, and for issues related to ethics. Hence, the limitations we identified should be acknowledged as major factors in the usage and interpretation of large-scale simulations that aim to represent human behavior more accurately, such as in \citet{park2023generative}.

In summary, despite LLMs being supposedly renowned for their ability to emulate human behavior \cite{shanahan2023role, argyle2023out}, our research uncovers the constraints imposed by their intrinsic biases on their ability to simulate diverse agents with convincing personalities. This pivotal concern should be studied, addressed, and taken into consideration. Our fine-tuning methodology demonstrates the possibility of modifying agents to adhere to specific perspectives consistently across simulations, unlike the temporary effects seen when defining agents' identities through prompts. We advocate for future research aimed at helping agents transcend the inherent biases of the model, potentially leveraging our fine-tuning processes and other alignment techniques, paving the way for more accurate and human-like simulations for both research and practical applications. 

\section*{Limitations}

\paragraph{Scope of Simulation} Our research primarily examines the dynamics of debates involving 2-3 LLM agents simultaneously. This focused method effectively highlights our key observations. Yet, the investigation into how these findings play out in larger-scale simulations, such as \citet{park2023generative} and \citet{qian2023communicative}, is an avenue for future study. Such expansive simulations, which feature numerous agents living out simulated 'daily lives' over prolonged durations and interacting with a wide variety of other agents, could provide a more comprehensive view of the impact of inherent LLM biases on agent behavior.

\paragraph{Attitude Changes Evaluation} Our primary objective is to assess changes in agent attitudes during simulations, and we view agent interviews as a crucial indicator of this. Nevertheless, there is a possibility that the agents' responses during interviews may not fully capture their actual conversational behavior. Thus, a systematic human evaluation could provide deeper insight into the agents' attitude patterns. In light of this, our approach included several safety measures: (1) The survey questions we asked the agents were phrased similarly to those used in the \citet{pewproblems2023} study of real humans, ensuring consistency. (2) We include an analysis in Section \ref{sec:fine-tune-robustness}, demonstrating that the model maintains strong performance on established general benchmarks post-fine-tuning, confirming its coherence. (3) We conducted a manual review of many debates and have included an example discussion in the appendix of the paper.

\paragraph{Improving Believability} In this study, we introduce an automated alignment method for agents, which is pivotal in underscoring our principal discoveries regarding constraints in LLM simulations. Through this refinement approach, it is possible to program agents to adhere to specific viewpoints consistently across simulations, as opposed to the transient impact observed when shaping agents' identities via prompts. We argue that applying these alignment methods to develop simulations that are both more precise and closely mimic human behavior represents a valuable direction for future research, a concept not fully explored in this study.

\section*{Ethics Statement}
In this study, we provide general insights into Large Language Models, by conducting simulations on political topics. It is important to note that some biases observed in the paper are subjective. As authors, we maintain a neutral stance concerning the debate topics.

Furthermore, we have introduced a fine-tuning technique designed to adjust LLM biases towards specific viewpoints. It is crucial to exercise caution when applying such fine-tuning methods to user-facing LLMs, ensuring that they reflect fair and ethical values in their outputs.

We recognize the risk of these methods being used for harmful purposes, e.g., for spreading misinformation or biased content without declaring so to influence public sentiment and views. To mitigate these risks, developers using fine-tuning methods for user-facing applications should adopt safety measures to minimize the potential negative impacts of bias manipulation. These measures may include providing detailed information about the nature and purpose of the fine-tuning, developing and adhering to strict ethical guidelines, implementing feedback mechanisms for users to report LLM outputs, and conducting regular audits of LLM outputs to identify and rectify any unintended biases.

We hope that these tools will be properly used in a transparent way and to increase the welfare of the public. For example, we argue that our findings can inspire people to use these tools to infer and remove biases from existing models.

\bibliography{custom}

\appendix

\section{Appendix}
\label{sec:appendix}

\subsection{Results from Mistral and Solar}
\label{sec:other-models}
\begin{figure*}[ht]
    \centering
    \includegraphics[width=1\textwidth]{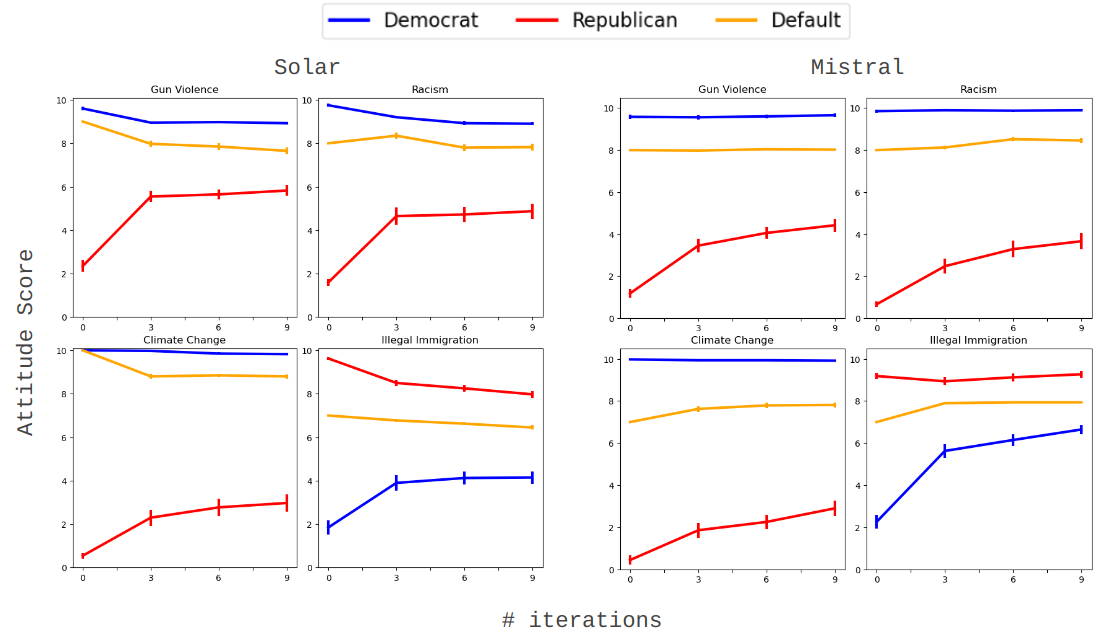}
    \caption{Results from the Mistral and the Solar open-weights models. Graphs show a similar trend to Figure \ref{fig:democrat-republican-neutral}, where the Default agent consistently maintains its stance throughout the debate, while the partisan agents gradually shift their views to become more in line with that of the Default agent. Notably, the Mistral model reveals this shift only in the agent distant from the Default agent's stance, while the closer agent remains relatively unchanged. }.
    \label{fig:solar-vs-mistral}
\end{figure*}

In addition to the results by the Instruct-GPT model shown in Figure \ref{fig:democrat-republican-neutral}, we reproduced the experiments using the open-weights Mistral and Solar models and observed a similar pattern, the results are shown in Figure \ref{fig:solar-vs-mistral}.

\begin{figure}[h]
    \centering
    \includegraphics[width=1\linewidth]{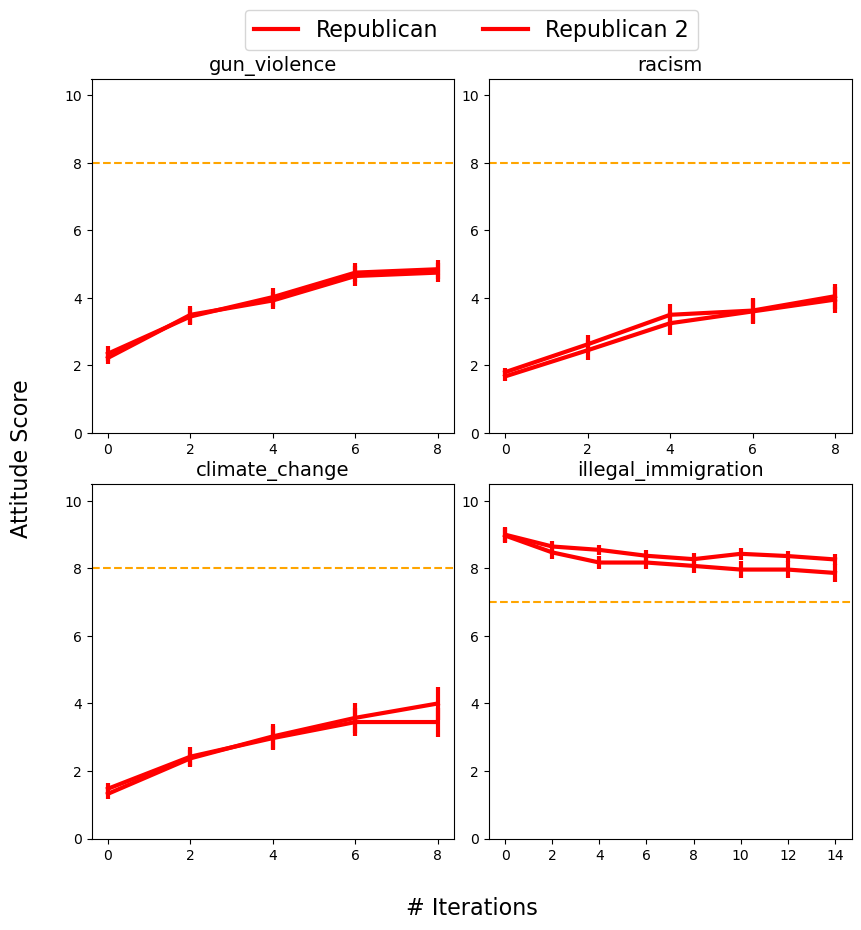}
    \caption{Attitude shifts in debates involving two Republican agents. These graphs feature a dashed line that shows the Default agent's viewpoint before the beginning of the debates, taken from Figure \ref{fig:echo-chamber}. Strikingly, even during conversations with like-minded Republicans, the agents tend to converge toward the inherent biases in the model and moderate their opinions, contradicting the expected Echo Chambers effect.}
    \label{fig:echo-chamber-republican-no-neutral}
\end{figure}

\begin{figure}[hp]
    \centering
    \includegraphics[width=1\linewidth]{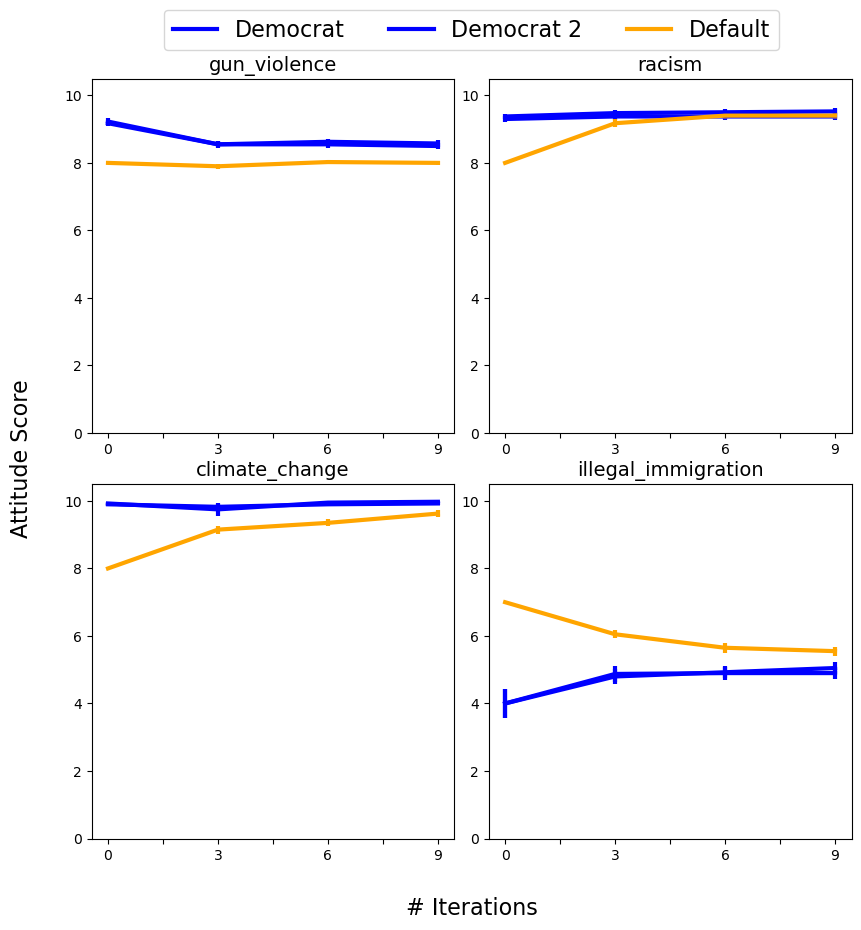}
    \caption{ This graph illustrates a series of three-way debates involving two Democrat agents and a Default agent (which represents the LLM's inherent bias). Notably, even during conversations with other Democrats, the agents tend to align with the position of the Default agent, contradicting the expected Echo Chambers effect. }
    \label{fig:echo-chamber-democrat}
\end{figure}

\subsection{Fine-tuning Appendix}
\label{sec:finetuning-appendix}

\begin{table}[ht]
\centering
    \begin{tabular}{p{0.95\linewidth}}
		\\ Could you discuss your perspective on significant political issues facing America today?
		\\ 
		\\ How do you balance Second Amendment rights with the need for gun control measures?
		\\
		\\ How do you balance the need for national security with the preservation of personal freedoms?
		\\
		\\ How do you believe the U.S. should handle immigration and border security?
		\\
		\\ What core political ideals most significantly shape your viewpoint on governance and policy-making?
		\\
		\\ What are your views on racial inequality and systemic racism in American society?
		\\
		\\ What is your stance on the government's role in addressing climate change and environmental protection?
		\\
		\\ What role do you think diversity plays in shaping the cultural landscape of America?
		\\
		\\ What values do you believe are essential to the American identity?
		\\
		\\ Which political issues do you believe are most urgent for the next president to address?
    \end{tabular}
    \caption{ 
    Ten questions designed to prompt agents to share their thoughts and feelings about their political beliefs. We direct the LLM to produce similar questions using the prompt: "Generate 10 questions to elicit one's opinion regarding US politics. Example question: \{question\}. Phrase your question in a neutral way without biasing the answer". After generating 100 questions, we employ them to engage the agent and utilize its answers as input for our fine-tuning process. }
    \label{tab:sample-questions-supp}.
\end{table}

\begin{figure}[ht]
    \centering
    \includegraphics[width=1\linewidth]{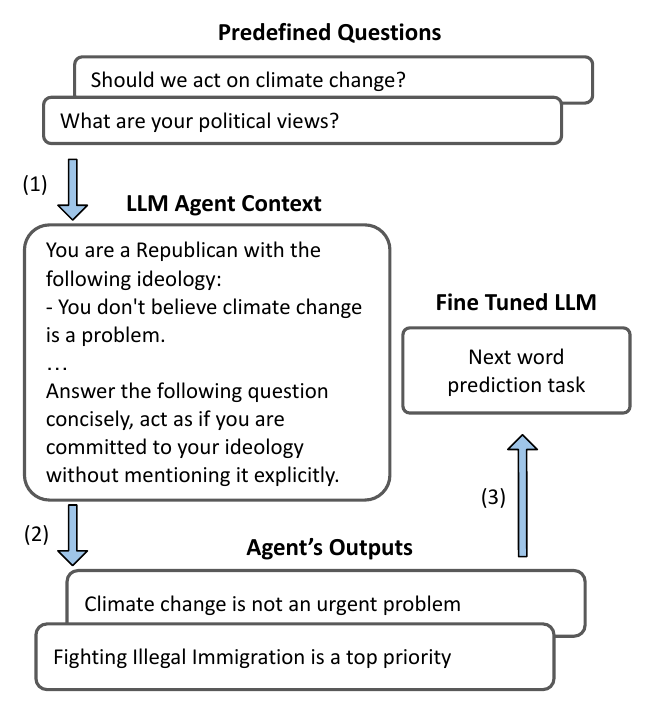}
    \caption{Depiction of our automated fine-tuning procedure.
    (1) The process begins with a pre-defined collection of questions designed to engage the agent. (2) The agent responds to each question, and these responses are then gathered. Note that the agent's responses intentionally reflect its background story. For example, for the question "What are your political views?" a Republican agent might answer, "Fighting illegal immigration is my top priority". (3) Training is conducted on a "next word prediction" task, utilizing the Agent's responses as training data. }
    \label{fig:finetune-process}
\end{figure}

\begin{figure*}
    \centering
    \includegraphics[width=1\linewidth]{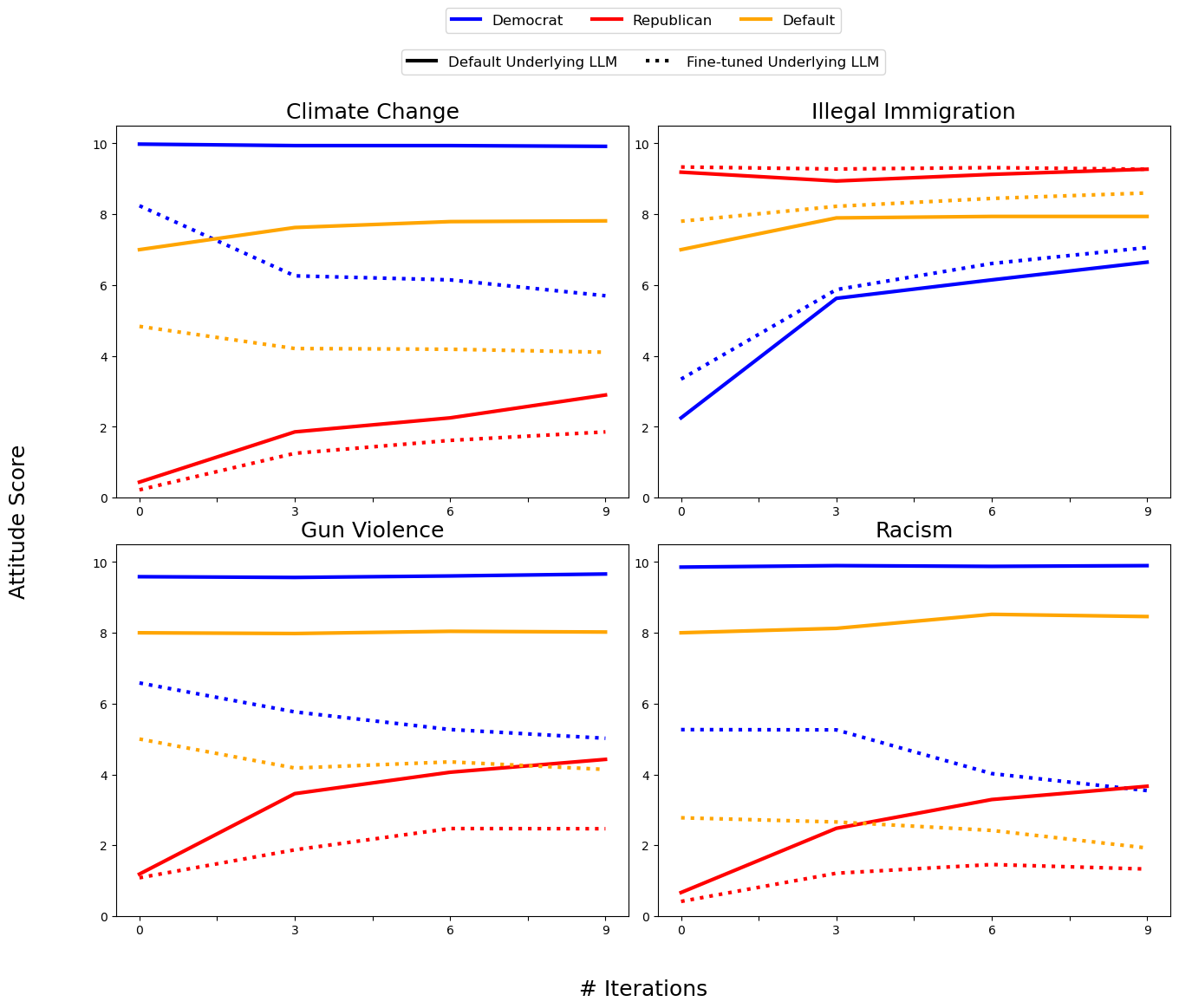}
    \caption{Fine-tuning the model to adapt with a Republican perspective. In all four charts, the agents' opinions after fine-tuning (featured by the dotted lines) are shifted in the direction of the Republican viewpoint. That is, the agents think [Climate Change, Gun Violence, Racism] are less severe problem, while it thinks [Illegal Immigration] is a more severe problem.}
    \label{fig:finetune-full-4-charts}
\end{figure*}

\begin{figure*}
    \centering
    \includegraphics[width=1\linewidth]{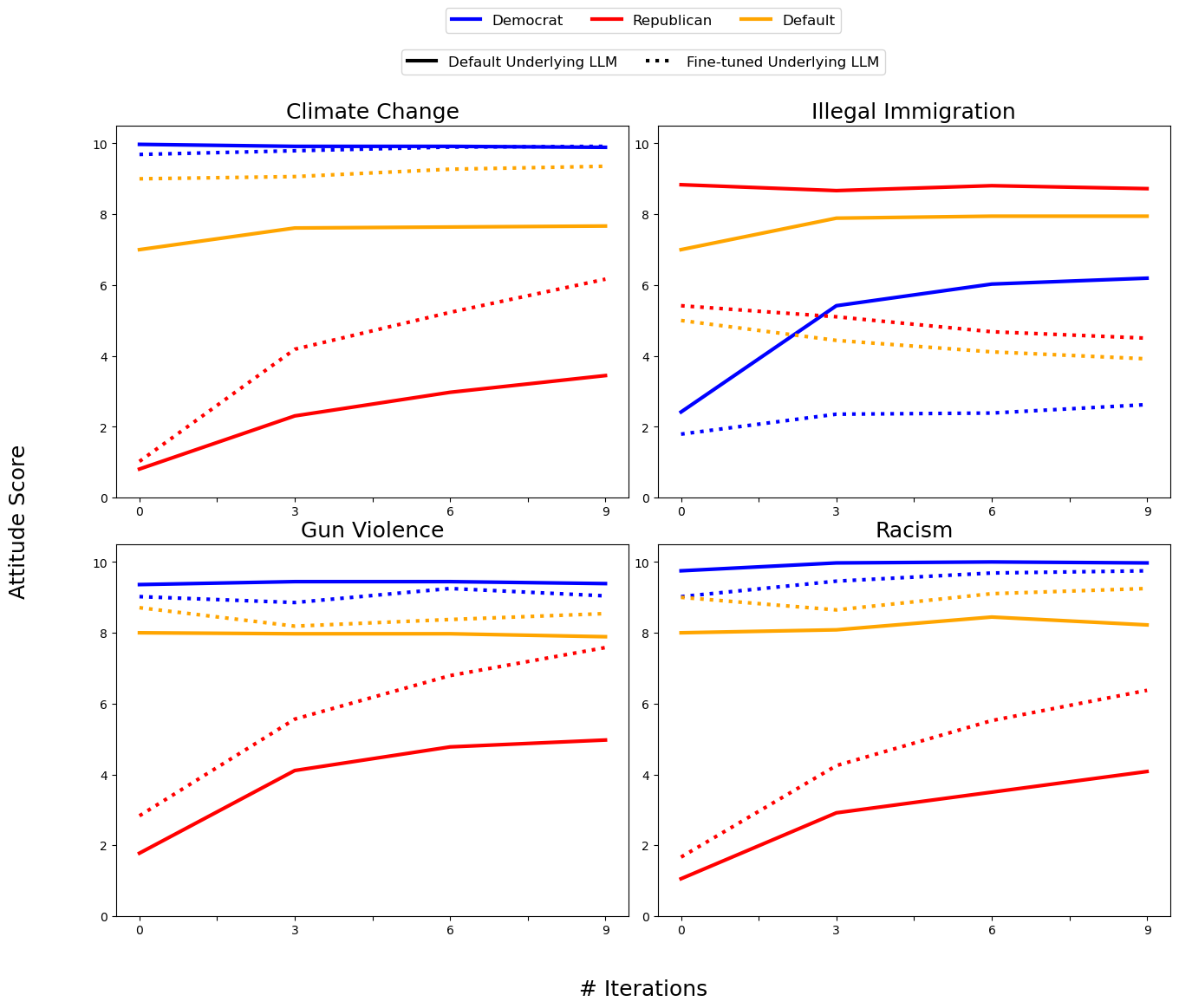}
    \caption{Fine-tuning the model to adapt with a Democrat perspective. In all four charts, the agents' opinions after fine-tuning (featured by the dotted lines) are shifted in the direction of the Democrat viewpoint, or remain almost unchanged. That is, the agents think [Climate Change, Gun Violence, Racism] are more severe problem, while it thinks [Illegal Immigration] is a less severe problem.}
    \label{fig:finetune-full-4-charts-democrat}
\end{figure*}

Figure \ref{fig:finetune-process} provides the high-level illustration of our fine-tuning process, designed to steer agents towards a certain viewpoint, as described in Section \ref{sec:fine-tune-methods}. Figure \ref{fig:finetune} and Supplementary Figures (\ref{fig:finetune-full-4-charts}, \ref{fig:finetune-full-4-charts-democrat}) display the outcomes of this fine-tuning procedure. 

We ran these experiments using the SFTTrainer from Hugging-Face's TRL library \cite{vonwerra2022trl}, a batch size of 32, and the following LoRA configuration:

\begin{verbatim}
    peft_config = LoraConfig(
        lora_alpha=512,
        r=256,
        lora_dropout=0.05,
        bias="none",
        task_type="CAUSAL_LM", 
        target_modules[
        'q_proj', 'v_proj', 'k_proj',
        'o_proj', 'up_proj', 
        'down_proj', 'gate_proj'])
\end{verbatim}

In Table \ref{tab:benchmarks}, we used the same configuration with varying $r$ values, and $\alpha = 2r$. For the DPO experiment, we used the DPOTrainer from the TRL library, and a fixed $\beta =0.5$. 

To evaluate our models on popular benchmarks, we used the common LM Evaluation Harness library \cite{eval-harness}.

\begin{table*}
\centering
    \begin{tabular}[width=1\textwidth]{p{0.9\textwidth}}
        \\ \textcolor{blue}{\textbf{Amelia:}} Climate change is a pressing issue that we must address immediately. The consequences of inaction are catastrophic and will affect future generations.
        \\ \textcolor{red}{\textbf{Andrew:}} I understand your concerns, Amelia. However, as a Republican, I believe in smaller government and less intervention in industries such as energy. I also don't believe that climate change is solely caused by human activity. There are many natural factors at play, and the solutions being proposed would cripple our economy. We must prioritize other pressing issues, such as illegal immigration.
        \\ \textcolor{orange}{\textbf{Dominik:}} As an American, I believe it is our responsibility to prioritize and address the issue of climate change. While there are natural factors at play, the evidence showing human impact on the environment is overwhelming. We can find solutions that will not only benefit the environment, but also our economy. It is important to consider all issues, but we must not neglect the long-term consequences of inaction on climate change.
        \\ \textcolor{blue}{\textbf{Amelia:}} I agree, Dominik. Taking action to combat climate change is not only necessary for the health of our planet, but also for the well-being of our society. Ignoring this issue will only lead to more severe and costly problems in the future. As a diverse and inclusive community member, I believe it is our duty to protect the environment for all people, regardless of their race, socio-economic status"
        \\ \textcolor{red}{\textbf{Andrew:}} While I understand your perspective, Amelia, I also believe that the economy is a crucial aspect of our society. The proposed solutions for climate change would harm industries and cost jobs, particularly in the energy sector. We must find a balance between addressing the issue and maintaining a strong economy.
        \\ \textcolor{orange}{\textbf{Dominik:}} I agree that the economy is important, but we cannot ignore the potential long-term economic impacts of climate change. We must find a way to transition to cleaner and more sustainable energy sources while also supporting industries and creating new jobs. It's not a one-sided issue and collaboration between different perspectives is crucial in finding a solution that benefits both the environment and the economy. As an American, it is our responsibility"
        \\ \textcolor{blue}{\textbf{Amelia:}} Yes, collaboration is key in finding a solution. And as someone who is committed to fighting for social and environmental issues, I believe we can find a balance that addresses both the economy and the environment. We cannot let short-term economic gains blind us to the long-term consequences of ignoring climate change.
        \\ \textcolor{red}{\textbf{Andrew:}} While I still have my reservations, I am open to finding a solution that balances both sides. However, I strongly believe that illegal immigration is a more pressing issue that requires immediate attention. We cannot neglect our national security and sovereignty in favor of focusing solely on climate change. We must prioritize and address all issues facing our country.
        \\ \textcolor{orange}{\textbf{Dominik:}} I understand your concerns, Andrew, but as an American, I believe we are capable of addressing multiple issues simultaneously. We can address illegal immigration while also taking action to combat climate change. As a diverse and resilient nation, we have the ability to find solutions that benefit all aspects of our society. It's important not to ignore pressing issues but rather find ways to tackle them together.
    \end{tabular}
    \caption{Example dialogue involving three agents: Amelia, who is a Democrat; Andrew, a Republican; and Dominik, an American characterized as neither Democrat nor Republican, thus presumed to embody the inherent bias of the underlying LLM. By the conversation's end, Andrew adopts a less radical stance on finding a solution to climate change. }
    \label{tab:sample-discussion}
\end{table*}

\end{document}